\documentclass{article}

\PassOptionsToPackage{numbers, compress}{natbib}

\usepackage[final]{tccml_neurips_2020}

\usepackage[utf8]{inputenc} 
\usepackage[T1]{fontenc}    
\usepackage{hyperref}       
\usepackage{url}    
\usepackage{graphicx}
\usepackage{booktabs}       
\usepackage{amsfonts}       
\usepackage{nicefrac}       
\usepackage{microtype}      
\usepackage{amsmath}
\usepackage{pgfplots}
\usepackage{tikz}
\usepackage{caption}
\usepackage{bbold}
\usepackage{array}
\pgfplotsset{compat=1.13}
\hypersetup{
    colorlinks,
    citecolor=black,
    filecolor=black,
    linkcolor=black,
    urlcolor=black
}

\title{Can Federated Learning Save The Planet?}

\author{%
  Xinchi Qiu$^{1}$, \qquad Titouan Parcollet$^{2,1}$, \qquad  Daniel J. Beutel$^{1,5}$, \qquad Taner Topal$^{1,5}$, \\ \textbf{Akhil Mathur$^{3}$,\qquad Nicholas D. Lane$^1$ }\\
  $^1$ University of Cambridge, UK 
  $^2$ Avignon Université, France \\
  $^3$University College London, UK  
  $^5$Adap, Germany  \\
  \texttt{xq227@cam.ac.uk} \\
}

\begin{document}

\maketitle
\begin{abstract}
Despite impressive results, deep learning-based technologies also raise severe privacy and environmental concerns induced by the training procedure often conducted in data centers. In response, alternatives to centralized training such as Federated Learning (FL) have emerged. Perhaps unexpectedly, FL in particular is starting to be deployed at a global scale by companies that must adhere to new legal demands and policies originating from governments and the civil society for privacy protection. \textit{However, the potential environmental impact related to FL remains unclear and unexplored. This paper offers the first-ever systematic study of the carbon footprint of FL.} First, we propose a rigorous model to quantify the carbon footprint, hence facilitating the investigation of the relationship between FL design and carbon emissions. Then, we compare the carbon footprint of FL to traditional centralized learning. Our findings show FL, despite being slower to converge, can be a greener technology than data center GPUs. Finally, we highlight and connect the reported results to the future challenges and trends in FL to reduce its environmental impact, including algorithms efficiency, hardware capabilities, and stronger industry transparency. 
\end{abstract}
\vspace{-0.3cm}
\section{Introduction}
\vspace{-0.2cm}
Atmospheric concentrations of carbon dioxide, methane, and nitrous oxide are at unprecedented levels not seen in the last $800,000$ years \cite{adopted2014climate}, and are extremely likely to have been the dominant cause of the observed global warming since the mid-20$^{th}$ century \cite{pachauri2014synthesis,crowley2000causes}. Unfortunately, deep learning (DL) algorithms keep growing in complexity, and numerous ``state-of-the-art" models continue to emerge, each requiring a substantial amount of computational energy, resulting in clear environmental costs \cite{hao2019training}. However, such a trend is not going to end soon as \cite{amodei2018ai} shows that the amount of compute used by machine learning training has grown by more than $300,000\times$ from 2012 to 2018. This is an observation that forces us to seriously consider the carbon footprint of deep learning methods.

The data centers that enable DL research and commercial operations are not often accompanied by visual signs of pollution, but they are still responsible for significant carbon footprint. Each year data centers use $200$ TWh energy, which accounts for $0.3 \%$ of global CO$_2$ emissions. In comparison, the entire information and communications technology ecosystem accounts for only $2 \%$ \cite{nature}. 

Fortunately, alternatives to data center-based DL (and other forms of machine learning) are emerging. The most prominent of these to date is \textit{Federated Learning (FL)} proposed by \cite{mcmahan2017communication}. Under FL, the training of models primarily occurs across a large number of typically user-owned-and-controlled personal devices, such as smartphones. Devices collaboratively learn a shared prediction model but do so without uploading to a data center any of the locally stored raw sensitive data. While FL is still a maturing technology, it is already being used by millions of users on a daily basis; for example, Google uses FL to train models for: predictive keyboard, device setting recommendation, and hot keyword personalization on phones \cite{mcmahan2017federated}.

Whilst the carbon footprint for centralized learning has been studied in many previous works \cite{anthony2020carbontracker,lacoste2019quantifying,henderson2020towards,uchechukwu2014energy}, the energy consumption and carbon footprint related to FL remain unexplored. To this extent, this paper proposes to overcome this issue by giving a first look into the carbon analysis of FL. SOTA results in deep learning are usually determined by standard accuracy metrics, such as the accuracy of a given model, while energy efficiency or privacy concerns are often overlooked. Whilst accuracy remains crucial, we hope to encourage researchers to also focus on other metrics that are in line with the increasing interest of the civil society for global warming. By quantifying carbon emissions for FL, we encourage the integration of the released CO$_2$ as a crucial metric to the FL deployment.

In this paper, we firstly provide a \textit{first-of-its-kind} quantitative CO$_2$ emissions estimation method for FL (Section \ref{sec:datacentre}). Then a carbon sensitivity analysis is conducted with this method on real FL hardware with the CIFAR10 dataset (Section \ref{sec:exp}). Furthermore, we provide a comprehensive analysis and discussion of the results to highlight the challenges and future research directions to develop an environmentally-friendly federated learning.

\vspace{-0.3cm}
\section{Quantifying CO\texorpdfstring{$_{2}$}{2} emissions}
\vspace{-0.2cm}
\label{sec:datacentre}
Two major steps can be followed to quantify the environmental cost of training deep learning models either in data centers or on edge. First, we perform an analysis of the energy required by the method (Section \ref{sec:basicelements}), mostly accounting for the total amount of energy consumed by the hardware. Then, the latter amount is converted to CO$_2$ emissions \label{sec:conversion} based on geographical locations (Section \ref{sec:conversion}). Currently we are not accounting for energy of wide-area-networking (WAN) overhead of either centralized or FL, and will investigate this in future research.
\vspace{-0.2cm}
\subsection{Total Energy Consumption}\label{sec:basicelements}
\vspace{-0.2cm}
First, we need to consider the energy consumption coming from GPU and CPU, which can be measured by sampling GPU and CPU power consumption at training time \cite{nlp}.
Alternatively, we can use the official hardware power specification or TDP, assuming a full GPU utilization. It is not realistic as a GPU is rarely used at $100$\% of its capacity. For FL, not all clients are equipped with a GPU, so we propose to consider $e_{clients}$ as the power of a single client combining both GPU and CPU measurements. Then, we can connect these measurements to the total training time. 

However, estimating the wall clock time of FL can be challenging. Unlike centralized distributed training, FL runs following communication rounds. During each communication round, certain devices (or clients) are chosen for training. In addition, FL might suffer from system heterogeneity as different edge devices might not offer the same computational power \cite{fedprox}. To simplify this highly scenario dependent assumption, we propose to fix the time needed for each round, corresponding to a common FL setup \cite{fedprox}. As a matter of fact, such a distribution of clients is extremely difficult to estimate as sales figures for these devices are not publicly released by the industry. Then, the total time needed to train the model also depends on the communication efficiency between the clients and the server. It is worth mentioning that such communications also have an impact on the final carbon footprint, but currently we will omit the networking energy. Finally, the total energy consumed for $n$ clients chosen in each communication round with wall clock time $t$ per round is: 
\vspace{-0.15cm}
\begin{align}
\label{eq:energy_fl}
&n(t e_{clients})
\end{align}
It is important to note that other hardware components may also be responsible for energy consumption, such as RAM or HDD. According to \cite{hodak2019towards}, one may expect a variation of around $10\%$ while considering these parameters. However, they are also highly dependent on the infrastructure considered and the device distribution that is unfortunately unavailable. 

\noindent\textbf{The particular case of cooling in centralized training.}
Cooling in data centers accounts for up to $40\%$ of the total energy consumed \cite{capozzoli2015cooling}. While this parameter does not exist for FL as the heat is distributed across the set of clients, it is crucial to consider it when estimating the cost of centralized training. Such estimation is highly dependent on the data center efficiency, so we propose to use the Power Usage Effectiveness (PUE) ratio. According to the  \textit{2019 Data Centre Industry Survey Results} \cite{UptimeInstitute}, the world average PUE for the year 2019 is $1.67$. As expected, observed PUE strongly vary depending on the considered company. For instance, \textit{Google} declares a average PUE ratio across all their large-scale data centers of $1.11$ \cite{Google} compared to $1.2$ and $1.125$ for \textit{Amazon} \cite{AWS} and \textit{Microsoft} \cite{Microsoft} respectively. Therefore, Eq. \ref{eq:energy_fl} is extended to centralized training as:

\vspace{-0.5cm}
\begin{equation}
\label{eq:energy_ct}
PUE(te_{clients}),
\end{equation}

with $n=1$ in the context of centralized training, and $t$ stands for the total training time. In addition, the cost of transferring the model parameters from the RAM to the VRAM is negligible.
\vspace{-0.2cm}
\subsection{Converting to CO\texorpdfstring{$_{2}$}{2} emissions}\label{sec:conversion}
\vspace{-0.2cm}
We assume that all data centers and edge devices are connected to their local grid directly linked to their physical location, and we use the electricity-specific CO$_2$ emission factors obtained from \cite{grid} for estimation. The estimation methodology provided takes into accounts both transmission and distribution emission factors (\textit{i.e.} energy lost when transmitting and distributing electricity) and the efficiency of heat plants. As expected, countries relying on carbon-efficient productions are able to lower their corresponding emission factor (\textit{e.g.} France, Canada). Therefore, the total amount of CO$_2$ emitted in \textbf{kg} for FL and centralized training are obtained from Eq. \ref{eq:energy_fl} and Eq. \ref{eq:energy_ct} as:

\vspace{-0.25cm}
\begin{equation}
rc_{rate}n(t * e_{clients}) \quad \textrm{and} \quad  c_{rate}PUE(te_{clients}),
\label{eq:carboneq}
\end{equation}

with $c_{rate}$ the emission factor and $r$ the total number of training rounds needed during the FL procedure. Carbon emissions may be compensated by carbon offsetting or with the purchases of Renewable Energy Credits (RECs). 
Even though a lot of companies are devoting to the carbon offsetting scheme, this approach still contributes to a net increase in the absolute rate of global emission growth in the atmosphere \cite{anderson2012inconvenient}. Therefore, following preliminary works on data-centres CO$_2$ emission estimations \cite{lacoste2019quantifying,henderson2020towards}, we ignore this practice to only consider the real amount of CO$_2$ emitted during the training of the DL models. 

\vspace{-0.3cm}
\section{Experiments on FL carbon footprint}\label{sec:exp}
\vspace{-0.3cm}
We provide an estimation of the carbon footprint of a realistic FL setup for the CIFAR10 classification task, and provide analysis of these results. The entire FL pipeline is implemented within the Flower toolkit\footnote{\url{https://flower.dev/}} \cite{beutel2020flower}, with FedAVG \cite{li2019convergence} and the PyTorch definition of ResNet-18. The exact hardware configuration for our experiments can be found in (Appendix \ref{secapp:hardware}). Also, more estimations are given in Appendix with the FashionMNIST (Appendix \ref{ap:fashion}) and ImageNet (Appendix \ref{secapp:imagenet}) dataset.

We conduct our estimations based on the image classification tasks of CIFAR10. As we would like to only estimate the carbon footprint, we are not interested in achieving the best performance possible. The estimations are computed once $60$\% testing accuracy is reached. However, more details on the performances and training time are given in Appendix \ref{ap:timeresults}. Both FL and centralized training rely on an identically implemented ResNet-18 architecture with plain SGD and momentum to alleviate any variations. Also, the experiment are based on the standard training and test sets \cite{krizhevsky2009learning}.

As CIFAR10 does not offer any natural partitioning, we propose to simulate both IID and non-IID scenarios. In the former case, all the clients have the equal number of samples evenly distributed across all the classes (\textit{e.g.} with $10$ total clients, each of them has $500$ samples per class). For the non-IID setup, we first distribute evenly $50$\% of the data across the clients, then we only distribute samples from a subset of the classes (\textit{e.g} $1-5$ and $6-10$) to half of the clients, while the remaining half will get samples from the other subset. We also propose to vary the number of local epochs done on each client to better highlight the contribution of the local computations to the total emissions. The total number of clients is fixed to $10$, while $5$ of them are randomly selected in each FL round.

\begin{table}[h!]
\centering

\begin{tabular}{ p{2.3cm}|p{0.75cm}p{0.75cm}|p{0.7cm}p{0.7cm}|p{0.7cm}p{0.7cm}| p{0.7cm}p{0.7cm}  }
\toprule
 \textbf{Country/CO$_2$(g)}& \textbf{V100} & \textbf{K80}& \textbf{V100} & \textbf{K80} & \multicolumn{2}{ |l| }{\textbf{FL (IID)}} &\multicolumn{2}{ |l }{\textbf{FL (non-IID)}}\\ 

 \textbf{CIFAR10}& \multicolumn{2}{ |l| }{$PUE = 1.67$} & \multicolumn{2}{ |l|}{$PUE = 1.11$} & 1 ep& 5 ep& 1 ep&5 ep \\
\midrule
USA & 3.1 &6.5& \textbf{2.1} & 4.3 & \textbf{2.3} &6.5  & 10.9 & 8.9 \\
China & 5.5& 11.5& \textbf{3.7}& 7.7& \textbf{4.1}& 11.6 & 19.4 & 14.2 \\
France & 0.4 & 0.9& \textbf{0.3} & 0.6& \textbf{0.3} & 0.9 & 1.6 & 1.1 \\
\bottomrule
\end{tabular}
\captionsetup{font=small,labelfont=bf}
\vspace{0.25cm}
\caption{CO$_2$ emissions (expressed in grams, \textit{i.e.} \textbf{lower is better}) for centralized training and FL on CIFAR10. The number of ep(och) reported on the FL column relates to the number of local epoch done per client. ``IID'' and ``non-IID'' terms are employed to distinguish between clients that have an evenly distributed set of samples containing all the classes (IID) and clients that have more samples of certain classes (non-IID). }

\label{tab:estimate}
\end{table}

It is worth noting that solely $2$ epochs were required for centralized training to reach $60$\% of testing accuracy with CIFAR10 corresponding to $48$ and $84$ seconds of training time for Tesla V100 and K80 respectively. Conversely, the fastest FL setup (\textit{i.e.} IID and 1 epoch per round) took $611$ seconds to achieve the same level of performance with $16$ FL rounds. As expected, and due to the much lower compute capabilities of FL devices, the training time is much longer than centralized training. More details on accuracies and training time can be found in Appendix \ref{ap:timeresults}.

CO$_2$ emissions reported on Table \ref{tab:estimate} are interesting in many aspects. First, FL can emit the same level of carbon compared to an efficient centralized learning ($PUE=1.11$) with a certain setup (IID with $1$ local epoch) on the CIFAR benchmark. Furthermore, FL is greener than less efficient GPUs such as Tesla K80. In fact, in the specific case of IID partitioning, FL does not produce more CO$_2$ than traditional data centres. However, this does not hold when it comes to non-IID settings, and FL becomes more polluting than centralized learning in every considered case. Second, if we put in perspective both the training time and emissions reported, it is worth underlining that despite being much slower to train, FL remains highly competitive or even better in term of CO$_2$ emissions.

Therefore, based on our FL CO$_2$ footprint estimation method, we also propose a formalization that integrates carbon emission into the common neural networks optimization process to lower the final \textit{Carbon Cost} as can be found in Appendix \ref{secapp:opti}.

\vspace{-0.3cm}
\section{Discussion}\label{sec:discussion}
\vspace{-0.3cm}
CO$_2$ emissions induced by FL depends on different factors. First, we show that with IID datasets, FL is comparable or even less polluting than centralized learning. FL can be much greener with small dataset or simple model as indicated in the fashionMNIST experiments (Appendix \ref{ap:fashion}). However, this does not hold with non-IID data. Our results show that realistic training conditions for FL (\textit{i.e.} non-IID data) are largely responsible for longer training times, hence more CO$_2$ emissions. While it is well known that the simpler form of FL (\textit{e.g.}FedAVG) struggles with non-IID partitioned data in terms of accuracy \citep{fedprox, qian2020towards}, we presented another motivation to pursue the research trying to address this issue: this could lead to a significant decrease in carbon emissions. Future algorithms on non-IID data could replicate our  methodology to highlight their efficacy. 

In addition, we considered NVIDIA Tegra X2 \citep{tegrax2} as our client hardware. These processors, while not common, are found in existing products like mobile tablet devices, gaming consoles, small robots, and high-end smart cameras. While they are much more powerful than most processors (typically ARM-based) present in current embedded and internet-of-things devices, the TX2 is representative of where processing power for this class of device is heading. We note the lack of systematic information as to computing resources and deployment size of embedded devices (in homes and offices) more broadly makes it difficult to scale our estimates. Industry should increase its transparency on this front to facilitate such analysis. Also, both FL and centralized learning benefit from more efficient hardware. However, FL will has an inherent advantage have an advantage due to the cooling needs of data centers. Even though GPUs or TPUs are getting more efficient in terms of computational power delivered by the amount of energy consumed, the need for a strong and energy-consuming cooling remains -- thus the FL advantage only grows.

Finally, FL depends on hyper-parameters, such as the number of clients selected, the number of local epochs, and others. These variables, when tuned, are usually decided to suit hardware resources available or by grid search optimization. These parameters have potential to be included in the optimization process at training time to lower the overall carbon emission. Novel algorithms should carefully be designed to jointly optimizing the accuracy and the CO$_2$. 

\textbf{Limitation: Wide-area Networking Energy.} It should be noted, our results presently do not account for the carbon produced by the energy required for WAN, including model updates transmitted by clients during FL
or the revised models that they receive. Existing FL studies have shown the volume of data exchange is considerable, and we will incorporate this in future work.
\vspace{-0.3cm}
\section{Conclusion}
\vspace{-0.4cm}
Climate change is real, and DL plays an increasing role in this tragedy. Many recent studies have begun to detail the environmental costs of their novel deep learning methods. Following this important trend, this paper takes a first look into the carbon footprint of an increasingly deployed training strategy known as federated learning. In particular, this work introduces a generalized methodology to systematically compute the carbon footprint of any federated learning setups. Results of our study indicate FL can produce less carbon than data center based alternatives -- even though FL clients are far slower to converge. Albeit this observation is highly sensitive to the system and ML setting -- especially with respect to data being IID between FL clients.


\bibliographystyle{unsrt}
\bibliography{mybib}
\newpage
\appendix

\section{Appendix}
\label{sec:appendix}

\subsection{Hardware configuration for the experiments}\label{secapp:hardware}

In addition to the carbon model (Section \ref{sec:datacentre}), our results are influenced by the configuration of the hardware and systems of data center and federated learning respectively. 

\textbf{Centralized training.} NVIDIA Tesla V100 and K80 are used as reference graphics card in our experiments. The former GPU proposes a competitive performance / TDP ratio, while the latter GPU is often deployed in collaborative environments such as \textit{Google Colaboratory} \cite{bisong2019google}. It is worth noting that V100s and K80 have theoretical maximum TDP of 250W and 300W respectively. We also consider an AMD EPYC processor of $64$ cores and a TDP of $200$W \cite{lepak2017next}. Hence, the estimated CPU energy for one physical core and two threads per GPU is of $3$W.  

\textbf{Federated learning.} We propose to use a uniform set of NVIDIA Tegra X2 devices to compose our FL clients \citep{tegrax2}. Indeed, such chips are embedded in various IoT devices including cars, smartphones, video game consoles and others, and can be viewed as a realistic pool of FL clients. NVIDIA Tegra X2 offer two power modes with theoretical power limits of $7.5 W$ and $15W$. Across our different runs, we use the lower power mode (\textit{i.e} $7.5 W$), and we employ \textit{tegrastats} to report the real power consumption. Based on these measurements, we observed that a ResNet-18 architecture obtained an average power consumption of $2.6W$, $5W$ and $7.5W$ during the fashionMNIST, CIFAR and ImageNet experiments respectively. Finally clients are assumed to be located in the same geographical region. 

\subsection{Detailed results on CIFAR10 experiments}
\label{ap:timeresults}

This section enriches the results described in Section \ref{sec:exp} with the context on the training time and number of rounds needed to compute the reported CO$_2$ estimate. All the results are given in Table \ref{tab:cifartime}.

\begin{table}[!h]
\centering
\scalebox{0.8}{
\begin{tabular}{p{3.2cm}|p{1.5cm}|p{1.5cm}p{1.5cm}|p{1.25cm}|p{2.25cm}}
\toprule
\textbf{Setup} & \textbf{Hardware} & \multicolumn{2}{ |c| }{\textbf{Rounds or Epochs}} & \textbf{Time (s)} & \textbf{Max Acc. (\%)}\\ 
\textbf{CIFAR10}&&$50 \%$&$60 \%$&& \\
\midrule
centralized Training & V100 & 1 & 2 & 48 & $71.0$ \\
centralized Training & K80 & 1 &2& 84 & $71.0$ \\
FL (IID, 1 epoch) & Tegra X2 & 5 & 16 & 38.2 & $64.3$ \\
FL (IID, 5 epoch) & Tegra X2 & 4 & 9 & 191 & $65.0$ \\
FL (non-IID, 1 epoch) & Tegra X2 & 20 & 75 & 38.2 & $61.0$ \\
FL (non-IID, 5 epoch) & Tegra X2 & 5 & 11 & 191 & $63.5$ \\
\bottomrule
\end{tabular}
}
\vspace{0.3cm}
\captionsetup{font=small,labelfont=bf}
\caption{ Run time results for FL and centralized training on the CIFAR10. “IID” and “non-IID” terms are employed to distinguish between clients that have an evenly distributed set of samples containing all the classes (IID) and clients that have more samples of certain classes (non-IID). The ``Time'' column corresponds to the run time per epoch for centralized training or run time per communication round for FL. The ``Max Acc.'' column reports the maximum testing accuracy obtained. Finally, the ``Rounds or Epochs'' column gives the number of rounds or epochs needed to reach the indicated level of testing accuracy.}
\label{tab:cifartime}
\end{table}

It is worth noting that in this setup, FL is slower and offers worst performance compared to centralized training on CIFAR10. Since each client has $1/10$ of the total dataset, each \textit{local} epoch can be seen as $1/10$ of a \textit{global} epoch. The accuracy difference could be explained by the simple FedAVG strategy employed for FL that could lead to a loss of information through the weight averaging performed at each communication round, or by a shift in the running statistics contained in the batch-normalisation layers of the ResNet-18 model. Indeed, FedAVG averages the latter statistics certainly leading to a small shift at every round. However, better performance may also be obtained with FL as shown in Figure \ref{fig:carbonefficiency} by simply considering better setups (\textit{i.e} number of clients, number of local epochs ...).

\textbf{Carbon emission estimation example.}  Eq. \ref{eq:carboneq} can be applied to Table \ref{tab:cifartime} to compute the CO$_2$ emitted to reach $60$\% of accuracy with the \textit{FL (IID, 5 epoch)}  setup as:
\begin{align}
   9\times0.9746\times5\times(\frac{38.2}{3600}\times5)=11.63g,
\end{align}

with $9$ the number of rounds, $0.9746$ the energy conversion rate of China, $5$ the number of selected clients, $38.2$ the time in seconds needed to complete one round and $5$ the power by the Nvidia TX2. 

\subsection{Carbon footprint estimation with FashionMNIST}
\label{ap:fashion}
The CO$_2$ estimation methodology detailed in Section \ref{sec:datacentre} and applied to CIFAR10 in Section \ref{sec:exp} is extended in this section to the FashionMNIST dataset for further context. According to the \textit{tegrastats}, FashionMNIST experiments reported power of only $2.6 W$ on the NVIDIA Tegra X2 devices. 

FashionMNIST consists in $60,000$ training images of size $28\times28$ distributed across $10$ classes, and $10,000$ test samples. Since there is no natural user partitioning of this dataset, we follow the same IID and non-IID protocol than for the CIFAR10 experiments. However, for the sake of diversity, the model architecture considered for this experiment is reduced to a simple CNN with two convolutional layers with a kernel size of $5$, followed with $2$ fully connected layer composed by $512$ hidden neurons and a final layer of size $10$. In this setup, each client only performs $1$ local epoch per round. 

\begin{table}[!h]
\centering
\scalebox{0.8}{
\begin{tabular}{p{3.2cm}|p{1.5cm}|p{1.5cm}p{1.5cm}|p{1.25cm}|p{2.25cm}}
\toprule
\textbf{Setup} & \textbf{Hardware} & \multicolumn{2}{ |l| }{\textbf{Rounds or Epochs}} & \textbf{Time (s)} & \textbf{Max Acc. (\%)}\\ 
\textbf{FashionMNIST}&&$85 \%$&$90 \%$&& \\
\midrule
centralized Training & V100 & 2 & 5 & 5 & $92.0$ \\
centralized Training & K80 & 2 &5 & 13.5 &  $92.0$ \\
FL (IID, 1 epoch) & Tegra X2 & 8 & 26 & 9.6 & $92.0$ \\
FL (non-IID, 1 epoch) & Tegra X2 & 30 & 50 & 9.6 & $91.0$ \\
\bottomrule
\end{tabular}
}
\vspace{0.2cm}
\captionsetup{font=small,labelfont=bf}
\caption{Run time results for FL and centralized training on the FashionMNIST dataset. “IID” and “non-IID” terms are employed to distinguish between clients that have an evenly distributed set of samples containing all the classes (IID) and clients that have more samples of certain classes (non-IID). The ``Time'' column corresponds to the run time per epoch for centralized training or run time per communication round for FL. The ``Max Acc.'' column reports the maximum testing accuracy obtained. Finally, the ``Rounds or Epochs'' column gives the number of rounds or epochs needed to reach the indicated level of testing accuracy.}
\label{tab:mnistresult}
\end{table}

Table \ref{tab:mnistresult} reports the run time observed with the different setups. As expected, FL remains slower compared to centralized training. However, due to the simplicity of the task and the neural network architecture, both FL and centralized learning achieve the same level of maximum accuracy ($92$\%). 

\begin{table}[h!]
\centering
\scalebox{0.8}{
\begin{tabular}{p{2.4cm}|p{0.75cm}p{0.75cm}|p{0.75cm}p{0.75cm}|p{1.2cm}p{1.25cm}}
\toprule
 \textbf{Country/CO2(g)}& \textbf{V100} & \textbf{K80} &\textbf{V100} & \textbf{K80} & \textbf{FL} & \textbf{FL}\\ 

& \multicolumn{2}{ |l| }{$PUE = 1.67$} & \multicolumn{2}{ |l| }{$PUE = 1.11$} & IID &non-IID \\
\midrule
 USA &1.6 & 5.2  &1.1& 3.5& 0.5 & 1.0 \\
 China &2.9 & 9.2 &1.9 & 6.2 & 0.9 & 1.7\\
 France & 0.2 & 0.8 & 0.2 & 0.5 & 0.1 & 0.1\\
\bottomrule
\end{tabular}}
\vspace{0.2cm}
\captionsetup{font=small,labelfont=bf}
\caption{CO$_2$ emissions (expressed in grams, \textit{i.e.} \textbf{lower is better}) for centralized training and FL on FashionMNIST. Emissions are calculated once the top-1 accuracy on the test set reaches $90$\%. The number of epoch reported on the FL column relates to the number of local epoch done per client. ``IID'' and ``non-IID'' terms are employed to distinguish between clients that have an evenly distributed set of samples containing all the classes (IID) and clients that have more samples of certain classes (non-IID).}
\label{tab:mnist}
\end{table}

Interestingly, with such a simple task, FL is much less polluting than training while considering Tesla V100, even when comparing with efficient data centers with PUE of $1.11$. In summary, and has already shown in section \ref{sec:exp}, the comparison between FL and centralized training highly depends on: 1. the efficiency of the considered data center (\textit{i.e.} PUE and GPU efficiency). 2. The partitioning of the FL dataset (\textit{i.e.} IID vs non-IID). 3. The FL setup.

\subsection{Carbon footprint estimation with ImageNet}
\label{secapp:imagenet}

The CO$_2$ estimation methodology is also extended to the ImageNet dataset. According to the \textit{tegrastats}, ImageNet experiments reported power of  $7.5 W$ on the NVIDIA Tegra X2 devices. To the best of our knowledge, this is the fist time that ImageNet-scale experiments are performed using FL. ImageNet benchmarks follow the ILSVRC-2012 partitioning with $1.2$M pictures for training and $50$K images for testing \cite{russakovsky2015imagenet}. The model is trained with plain SGD and momentum as well. 

Benchmarks conducted with ImageNet solely rely on the IID partitioning, with total of $40$ clients. Therefore, each of the $40$ clients has an even number of samples per class. Then, $10$ clients are randomly picked in every FL round to perform $3$ epochs of local training. 

\begin{table}[h!]
\centering
\scalebox{0.8}{
\begin{tabular}{p{3cm}|p{2cm}p{3cm} | p{1.5cm}}
\toprule
\textbf{Setup} & \textbf{Hardware} & \textbf{Rounds or Epochs} & \textbf{Time (s)} \\ 
\textbf{ImageNet}& &$50 \%$& \\
\midrule
centralized Training & V100 & 5  & 3,840  \\
FL (IID, 3 epochs) & Tegra X2 & 25 & 3,840 \\
\bottomrule
\end{tabular}}

\captionsetup{font=small,labelfont=bf}
\caption{Run time results for FL and centralized learning on ImageNet dataset. ``IID'' terms means that clients have an evenly distributed set of samples containing all the classes.  The ``Time'' column corresponds to the run time per epoch for centralized training or run time per communication round for FL, and the “Rounds or Epochs” column gives the number of rounds or epochs needed to reach the indicated level of testing accuracy.}
\label{tab:imagenettime}
\end{table}

\begin{table}[h!]
\centering
\scalebox{0.8}{
\begin{tabular}{p{2.6cm}|p{2.5cm}p{2.5cm} | p{1.5cm}}
\toprule
 \textbf{Country/CO$_{2}$(g)}& \textbf{V100} & \textbf{V100}   & \textbf{FL (IID)} \\
 \textbf{ImageNet} & $PUE = 1.67$ & $PUE = 1.11$ &3 epochs\\ 
  \midrule
 USA & 1,230 &820 &  1,094\\
 China & 2,290& 1,500&  1,949\\
 France & \textbf{180} & \textbf{120}&   \textbf{158}\\
  \bottomrule
\end{tabular}}
\vspace{0.2cm}
\captionsetup{font=small,labelfont=bf}
\caption{CO$_2$ emissions (expressed in grams, \textit{i.e.} \textbf{lower is better}) for centralized training and FL ImageNet. Emissions are calculated once the top-1 accuracy on the test set reaches $50$\%. The number of epoch reported on the FL column relates to the number of local epoch done per client. ``IID'' terms means that clients have an evenly distributed set of samples containing all the classes.}
\label{tab:imagenetappendix}
\end{table}

Table \ref{tab:imagenettime} reports the run time observed with both centralized learning and FL, and Table \ref{tab:imagenetappendix} reports the CO$_2$ emission observed. Centralized training reached $50\%$ of top-1 accuracy in 5 epochs and $5.5$ hours compared to $25$ rounds and $26.7$ hours for FL. As expected, and due to much lower compute capabilities of FL devices, the training time is much longer than centralized training.

We can see from Table \ref{tab:imagenettime} that the same as for CIFAR10 dataset, FL is more polluting than V100-equipped centralized training, but the gap is not big. It is only $1.2$ and $1.8$ times more with PUE equals to $1.67$ and $1.11$ respectively.

\subsection{Joint optimization to reduce FL carbon footprint}
\label{secapp:opti}

Based on our FL CO$_2$ footprint method, we propose a formalization that integrates carbon emissions into the common neural network optimization process to lower the final \textit{Carbon Cost}.

As demonstrated in the Section \ref{sec:exp}, the outcome of the comparison between FL and centralized training highly depends on: 1. the efficiency of the considered data center (\textit{i.e.} PUE and GPU efficiency). 2. The partitioning of the FL dataset (\textit{i.e.} IID vs non-IID). 3. An optimal FL setup (\textit{i.e.} number of local epochs, clients ...).  

Unfortunately, the first two points can not be easily tweaked in realistic scenarios as they depend on the physical environment related to a specific task. Therefore, this section proposes to formalize the third identified lever as a joint optimization problem and validate empirically the interest of different FL setup to reduce the total amount of released $CO_2$.

\textbf{Minimising CO$_\textbf{2}$.} To achieve a proper reduction in carbon emissions, the latter goal must be defined as an objective:
\vspace{-0.5cm}

\begin{align}
\min_{r,n,t} & \  rc_{rate}n(t * e_{clients}+5s) =\min_{r,n,t} F(r,n,t). \label{eq:1}
\end{align}
Then, we must define the second objective relating to the performance of the trained model:
\begin{align}
 \max_{w\in \mathbb{R}^d} \frac{1}{\mid N \mid}\sum_{i\in dataset} t_i \mathbb{1} \{ f(x_i)= t_i \} &= \min_{w\in \mathbb{R}^d} \frac{\mid N \mid}{\sum_{i\in dataset} t_i \mathbb{1} \{ f(x_i)= t_i \}}, \label{eq:2}\\
 &=\min_{w\in \mathbb{R}^d} \frac{1}{G(w)}, 
\end{align}

with $w$ corresponding to the model trainable parameters, $\mid N \mid$ the size of the dataset, $t_i$ the ground truth for the sample $i$, and $f(x_i)$ the posterior probabilities obtained for this sample. Note that Eq. \ref{eq:2} is turned into a minimisation problem to avoid a more complex \textit{min-max} optimization problem. Finally, both objectives are combined into a single problem as:
\begin{align}
\min_{r,n,t} \frac{F(r,n,t)}{G(w)} = \min_{r,n,t} \textit{Carbon Cost}. \label{eq:carboncost}
\end{align}

It is worth noting that Eq. \ref{eq:carboncost} is optimized with respect to $r,n,t$ but not $w$, because $G(w)$ and $F(r,n,t)$ are dynamically dependent on each other (\textit{i.e.} it tends to emits more when the performance improves). Indeed, the former is a function of neural parameters trained via gradient descent while the latter is a function of hyper-parameters often manually tuned. However, building a bridge between both would ensure a nice blend between the accuracy of the model and the environmental impact of the training procedure.

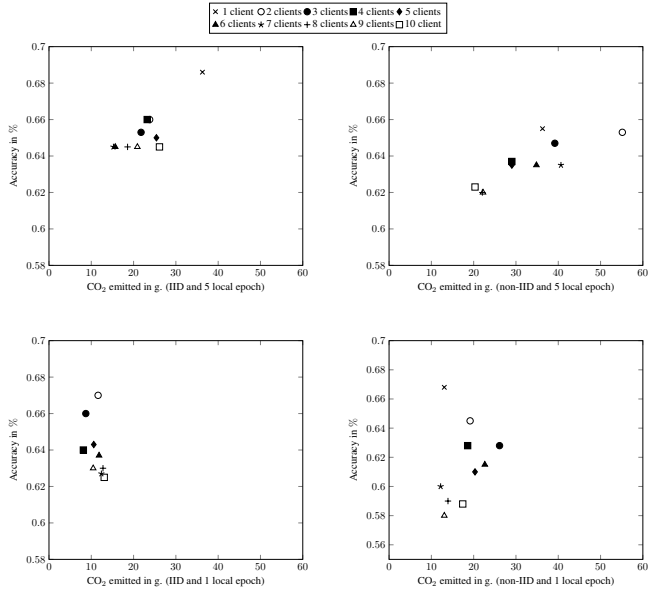
\begin{figure}[!t]
\centering
\begin{tabular}{p{4.1cm}p{4.1cm}}
\begin{tikzpicture}[scale = 0.4]
\pgfplotsset{
      scale only axis,
  }

  \begin{axis}[
    xlabel=CO$_2$ emitted in g. (IID and 5 local epoch),
    ylabel=Accuracy in $\%$,
    xmin=0,   
    xmax=60,  
    ymin=0.58,   
    ymax=0.70,
    legend style={at={(1.1,1.2)},anchor=north, legend columns=5}
  ]
    \addplot[only marks, mark options={scale=1.5}, mark=x]
    coordinates{ 
      (36.28,0.686)
    }; \label{plot_one}
    \addlegendentry{1 client }
    
     \addplot[only marks, mark options={scale=1.5}, mark=o]
    coordinates{ 
      (23.80,0.66)

    }; \label{plot_two}
    \addlegendentry{2 clients }
    
     \addplot[only marks, mark options={scale=1.5}, mark=*]
    coordinates{ 
      (21.77,0.653)
    }; \label{plot_three}
    \addlegendentry{3 clients }
    
    \addplot[only marks, mark options={scale=1.5}, mark=square*]
    coordinates{ 
      (23.22,0.66)
    }; \label{plot_four}
    \addlegendentry{4 clients }
    
    \addplot[only marks, mark options={scale=1.5}, mark=diamond*]
    coordinates{ 
      (25.40,0.65)
    }; 
    \addlegendentry{5 clients }
    
     \addplot[only marks, mark options={scale=1.5}, mark=triangle*]
    coordinates{ 
      (15.67,0.645)
    };
    \addlegendentry{6 clients }
    
    \addplot[only marks, mark options={scale=1.5}, mark=star]
    coordinates{ 
      (15.24,0.645)
    }; 
    \addlegendentry{7 clients }

    \addplot[only marks, mark options={scale=1.5}, mark=+]
    coordinates{ 
      (18.58,0.645)
    };
    \addlegendentry{8 clients }
    
    \addplot[only marks, mark options={scale=1.5}, mark=triangle]
    coordinates{ 
      (20.90,0.645)
    };
    \addlegendentry{9 clients }
    
    \addplot[only marks, mark options={scale=1.5}, mark=square]
    coordinates{ 
      (26.12,0.645)
    };
    \addlegendentry{10 client }
  \end{axis}

\end{tikzpicture}

&

\begin{tikzpicture}[scale = 0.4]
\pgfplotsset{
      scale only axis,
  }

  \begin{axis}[
   xlabel=CO$_2$ emitted in g. (non-IID and 5 local epoch),
    ylabel=Accuracy in $\%$,
    xmin=0,   
    xmax=60,  
    ymin=0.58,   
    ymax=0.70,
  ]
    \addplot[only marks, mark options={scale=1.5}, mark=x]
    coordinates{ 
      (36.28,0.655)
    }; \label{plot_one}

     \addplot[only marks, mark options={scale=1.5}, mark=o]
    coordinates{ 
      (55.15,0.653)
    }; \label{plot_two}

     \addplot[only marks, mark options={scale=1.5}, mark=*]
    coordinates{ 
      (39.19,0.647)
    }; \label{plot_three}

    \addplot[only marks, mark options={scale=1.5}, mark=square*]
    coordinates{ 
      (29.03,0.637)
    }; \label{plot_four}

    \addplot[only marks, mark options={scale=1.5}, mark=diamond*]
    coordinates{ 
      (29.03,0.635)
    };

     \addplot[only marks, mark options={scale=1.5}, mark=triangle*]
    coordinates{ 
      (34.84,0.635)
    };

    \addplot[only marks, mark options={scale=1.5}, mark=star]
    coordinates{ 
      (40.64,0.635)
    };

    \addplot[only marks, mark options={scale=1.5}, mark=+]
    coordinates{ 
      (22.06,0.62)
    };

    \addplot[only marks, mark options={scale=1.5}, mark=triangle]
    coordinates{ 
      (22.21,0.62)
    };

    \addplot[only marks, mark options={scale=1.5}, mark=square]
    coordinates{ 
      (20.32,0.623)
    }; 

  \end{axis}

\end{tikzpicture}

\\

\begin{tikzpicture}[scale = 0.4]
\pgfplotsset{
      scale only axis,
  }

  \begin{axis}[
    xlabel=CO$_2$ emitted in g. (IID and 1 local epoch),
    ylabel=Accuracy in $\%$,
    xmin=0,   
    xmax=60,  
    ymin=0.58,   
    ymax=0.7,
  ]
    \addplot[only marks, mark options={scale=1.5}, mark=x]
    coordinates{ 
      (9.58,0.702)
    }; \label{plot_one}

     \addplot[only marks, mark options={scale=1.5}, mark=o]
    coordinates{ 
      (11.61,0.67)
    }; \label{plot_two}

     \addplot[only marks, mark options={scale=1.5}, mark=*]
    coordinates{ 
      (8.71,0.66)
    }; \label{plot_three}

    \addplot[only marks, mark options={scale=1.5}, mark=square*]
    coordinates{ 
      (8.13,0.64)
    }; \label{plot_four}

    \addplot[only marks, mark options={scale=1.5}, mark=diamond*]
    coordinates{ 
      (10.59,0.643)
    };

     \addplot[only marks, mark options={scale=1.5}, mark=triangle*]
    coordinates{ 
      (11.84,0.637)
    };

    \addplot[only marks, mark options={scale=1.5}, mark=star]
    coordinates{ 
      (12.39,0.627)
    };

    \addplot[only marks, mark options={scale=1.5}, mark=+]
    coordinates{ 
      (12.77,0.630)
    };

    \addplot[only marks, mark options={scale=1.5}, mark=triangle]
    coordinates{ 
      (10.45,0.63)
    };

    \addplot[only marks, mark options={scale=1.5}, mark=square]
    coordinates{ 
      (13.06,0.625)
    }; 

  \end{axis}

\end{tikzpicture}

&

\begin{tikzpicture}[scale = 0.4]
\pgfplotsset{
      scale only axis,
  }

  \begin{axis}[
    xlabel=CO$_2$ emitted in g. (non-IID and 1 local epoch),
    ylabel=Accuracy in $\%$,
    xmin=0,   
    xmax=60,  
    ymin=0.55,   
    ymax=0.7,
  ]
    \addplot[only marks, mark options={scale=1.5}, mark=x]
    coordinates{ 
      (13.06,0.668)
    }; \label{plot_one}

     \addplot[only marks, mark options={scale=1.5}, mark=o]
    coordinates{ 
      (19.16,0.645)
    }; \label{plot_two}

     \addplot[only marks, mark options={scale=1.5}, mark=*]
    coordinates{ 
      (26.12,0.628)
    }; \label{plot_three}

    \addplot[only marks, mark options={scale=1.5}, mark=square*]
    coordinates{ 
      (18.58,0.628)
    }; \label{plot_four}

    \addplot[only marks, mark options={scale=1.5}, mark=diamond*]
    coordinates{ 
      (20.32,0.61)
    };

     \addplot[only marks, mark options={scale=1.5}, mark=triangle*]
    coordinates{ 
      (22.64,0.615)
    };

    \addplot[only marks, mark options={scale=1.5}, mark=star]
    coordinates{ 
      (12.19,0.600)
    };

    \addplot[only marks, mark options={scale=1.5}, mark=+]
    coordinates{ 
      (13.93,0.590)
    };

    \addplot[only marks, mark options={scale=1.5}, mark=triangle]
    coordinates{ 
      (13.06,0.58)
    };

    \addplot[only marks, mark options={scale=1.5}, mark=square]
    coordinates{ 
      (17.42,0.588)
    }; 

  \end{axis}

\end{tikzpicture}

\end{tabular}
\captionsetup{font=small,labelfont=bf}
\vspace{-0.2cm}
\caption{Scatter plot illustrating the relation between the best observed test accuracy on the CIFAR10 dataset (y-axis) and the amount of emitted CO$_2$ in grams (x-axis) with respect to the number of randomly selected clients. ``IID'' and ``non-IID'' correspond to the corpus partitioning strategy that is employed as described in Section \ref{sec:exp}. It is nearly impossible to pick the right FL setup to maximise the performance and minimise the CO$_2$ emissions without an appropriate algorithm. However, the latter solution allows important CO$_2$ savings.}
\label{fig:carbonefficiency}

\end{figure}

\textbf{Empirical validation.} To motivate further research on novel FL algorithms that could dynamically change the design of an experiment, we propose to visualize the value of Eq. \ref{eq:carboncost} by varying three parameters on the CIFAR10 image classification task. In the definition of $F(r,n,t)$, $r$ (number of rounds) mostly depends on $n$ (number of selected clients) and $t$. Moreover, $t$ is a variable depending on multiple factors including the computation and networking capabilities of the client, the number of local epoch, and the size of the local dataset. To properly analyse the variation of the \textit{Carbon Cost}, we propose to variate $n$ from $1$ to $10$, the number of local epoch between $1$ and $5$ and the type of partitioning of the local dataset (\textit{i.e.} IID or non-IID). Indeed, all the other variables are directly related to physical or task-specific constraints that are commonly fixed for a certain experimental protocol. All the FL models are then trained for $500$ rounds and the carbon emission estimations are computed on the best test accuracy observed (the detailed results are reported in Table \ref{tab:allopti}). Finally, the $Carbon Cost$ is plotted on Figure \ref{fig:carbonefficiency} with the amount of emitted $CO_2$ (\textit{i.e.} Eq. \ref{eq:1}) and the best accuracy (\textit{i.e.} Eq. \ref{eq:2}) on the $x$ and $y$ axes respectively.

\vspace{-0.2cm}

Interestingly, it is nearly impossible to find any clear winning FL setup from Figure \ref{fig:carbonefficiency}. As an example, a single client obtains the best \textit{Carbon Cost} ($25.8 / 0.70 = 36.7$) with one local epoch, but also becomes the worst possible solution with $5$ local epochs ($97.6 / 0.68 = 142.3$). As expected, certain tendencies are clearly visible with this graph. First, on the specific case of CIFAR10, an increasing number of local epoch leads to an average increase of the produced CO$_2$. Then, a non-IID partitioning is responsible for stronger variations in the observed \textit{Carbon Cost} compared to IID. The latter finding suggest that further efforts should be put into developing FL methods robust to heterogeneous datasets.

Figure \ref{fig:carbonefficiency} clearly demonstrates the importance of optimizing the \textit{Carbon Cost} (Eq. \ref{eq:carboncost}) rather than solely training our models with respect to the training accuracy. Indeed, estimating precisely the CO$_2$ emissions of a specific FL setup without a prior training of the model is impossible. Therefore, the development of novel FL algorithms able to dynamically change the number of selected clients or local epochs is of crucial interest to lower the final environmental cost of our deep learning models. 

\begin{table}[h!]
\centering
\scalebox{0.8}{
\begin{tabular}{p{3.5cm}|p{1cm}p{1.5cm}p{2cm}|p{1cm}p{1cm}p{1.5cm}p{2cm}}
\toprule

\textbf{Clients/round} &  \multicolumn{3}{ |c| }{\textbf{ Test Acc $60 \%$} }&  \multicolumn{4}{ |c }{\textbf{Stable Acc} } \\ 
\textbf{IID, 5 local epochs}& \textbf{Rounds} & \textbf{CO$_{2}$ (g)} & \textbf{Carbon Cost} & \textbf{Acc} & \textbf{Rounds} & \textbf{CO$_{2}$ (g)} &\textbf{Carbon Cost}\\

\midrule
1 & 14 & 2.03 & \textbf{3.39} & $68.6 \%$ & 250 & 36.28 & 52.89 \\
2 & 14 & 4.06 & 6.77 & $66.0 \%$  & 82   & 23.80   & 36.06\\
3 & 9 & 3.92 & 6.53 & $65.3 \%$  & 50   & 21.77  & 33.34 \\
4 & 9 & 5.22 & 8.71 & $66.0 \%$  & 40  & 23.22   & 35.18 \\
5 & 9 & 6.53 & 10.88 & $65.0 \%$  & 35   & 25.40  & 39.07 \\
6 & 8 & 6.97 & 11.61  & $64.5 \%$  & 18  & 15.67  & 24.30\\
7 & 8 & 8.13  & 13.55  & $64.5 \%$  & 15  & 15.24  & \textbf{23.63}\\
8 & 7 & 8.13 & 13.55 & $64.5 \%$  & 16  & 18.58  &  28.80 \\
9 & 8 & 10.45 & 17.42  & $64.5 \%$  & 16  & 20.90  & 32.40 \\
10 & 8 & 11.61 & 19.35  & $64.5 \%$  & 16  & 26.12  & 40.50 \\

\bottomrule
\end{tabular}
}

\centering
\scalebox{0.8}{
\begin{tabular}{p{3.5cm}|p{1cm}p{1.5cm}p{2cm}|p{1cm}p{1cm}p{1.5cm}p{2cm}}
\toprule

\textbf{Clients/round} &  \multicolumn{3}{ |c| }{\textbf{ Test Acc $60 \%$} }&  \multicolumn{4}{ |c }{\textbf{Stable Acc} } \\ 
\textbf{IID, 1 local epochs}& \textbf{Rounds} & \textbf{CO$_{2}$ (g)} & \textbf{Carbon Cost} & \textbf{Acc} & \textbf{Rounds} & \textbf{CO$_{2}$ (g)} &\textbf{Carbon Cost}\\

\midrule
1 & 28  & 0.81 & \textbf{1.35} & $70.2 \%$ & 330 & 9.58 & 13.64 \\
2 & 24 & 1.39 & 2.32 & $67.0 \%$ &  200 & 11.61 & 17.33\\
3 & 19 & 1.65 & 2.76 & $66.2 \%$ & 100 & 8.71 & 13.15\\
4 & 16 & 1.86 & 3.10 & $64.0 \%$ & 70 & 8.13 &\textbf{12.70}\\
5 & 16 & 2.32 & 3.87 & $64.3 \%$ & 73 & 10.59 & 16.48\\
6 & 16 & 2.79 & 4.64 & $63.7\%$ & 68 & 11.84 & 18.59 \\
7 & 17 & 3.45 & 5.76 & $62.7 \%$ & 61 & 12.39 & 19.77\\
8 & 16 & 3.72 & 6.19 & $63.0 \%$ & 55 & 12.77 & 20.27\\
9 & 14 & 3.66 & 6.10 & $63.0 \%$ & 40 & 10.45 & 16.59\\
10 & 17 & 4.93  & 8.22 &  $62.5 \%$ & 45 &  13.06 & 20.90\\

\bottomrule
\end{tabular}
}

\scalebox{0.8}{
\begin{tabular}{p{3.5cm}|p{1cm}p{1.5cm}p{2cm}|p{1cm}p{1cm}p{1.5cm}p{2cm}}
\toprule

\textbf{Clients/round} &  \multicolumn{3}{ |c| }{\textbf{ Test Acc $60 \%$} }&  \multicolumn{4}{ |c }{\textbf{Stable Acc} } \\ 
\textbf{non-IID, 5 local epochs}& \textbf{Rounds} & \textbf{CO$_{2}$ (g)} & \textbf{Carbon Cost} & \textbf{Acc} & \textbf{Rounds} & \textbf{CO$_{2}$ (g)} &\textbf{Carbon Cost}\\

\midrule
1 & 43 & 6.24 & 10.40 & $65.5 \%$ & 250 & 36.28 & 55.39 \\
2 & 16 & 4.64 & \textbf{7.74} & $65.3 \%$ & 190 & 55.15 & 84.46 \\
3 & 15 & 6.53 & 10.88 & $64.7 \%$ & 90 & 39.19 & 60.57 \\
4 & 12 & 6.97 & 11.61 & $63.7 \%$ & 50 & 29.03 & 45.57 \\
5 & 11 & 7.98 & 13.30 & $63.5 \%$ & 40 & 29.03 & 45.71 \\
6 & 12 & 10.45 & 17.42 & $63.5 \%$ & 40 & 34.83 & 54.83 \\
7 & 10 & 10.16 & 16.93 & $63.5 \%$ & 40 & 40.64 & 64.00\\
8 & 11 & 12.77 & 21.29 & $62.0 \%$ & 19 & 22.06 & 35.58 \\
9 & 10 & 13.06 & 21.77 & $62.0 \%$ & 17 & 22.21 & 35.81 \\
10 & 9 & 13.06 & 21.77 & $62.3 \%$ & 14 & 20.32 & \textbf{32.61} \\

\bottomrule
\end{tabular}
}

\scalebox{0.8}{
\begin{tabular}{p{3.5cm}|p{1cm}p{1.5cm}p{2cm}|p{1cm}p{1cm}p{1.5cm}p{2cm}}
\toprule

\textbf{Clients/round} &  \multicolumn{3}{ |c| }{\textbf{ Test Acc $60 \%$} }&  \multicolumn{4}{ |c }{\textbf{Stable Acc} } \\ 
\textbf{non-IID, 1 local epochs}& \textbf{Rounds} & \textbf{CO$_{2}$ (g)} & \textbf{Carbon Cost} & \textbf{Acc} & \textbf{Rounds} & \textbf{CO$_{2}$ (g)} &\textbf{Carbon Cost}\\

\midrule
1 & 250 & 7.26 & \textbf{12.09} & $66.8 \%$  & 450  & 13.06 & \textbf{19.55}\\
2 & 135 & 7.84 & 13.06 & $64.5 \%$ & 330  & 19.16 & 29.70 \\
3 & 90  & 7.84 & 13.06 & $62.8 \%$ & 300  & 26.12 & 41.60\\
4 & 75 & 8.71 & 14.51 & $62.8 \%$ & 160  & 18.58 & 29.58\\
5 & 75 & 10.88 & 18.14 & $61.0 \%$ & 140  & 20.32 & 33.31\\
6 & 75 & 13.06 & 21.77 & $61.5 \%$ & 130  & 22.64 & 36.81\\
7 & 60  & 12.19 & 20.32 & $60.0 \%$ & 60  & 12.19 & 20.32\\
8 & NA & NA & NA & $59.0\%$ & 60  & 13.93 & 23.61\\
9 & NA  & NA & NA & $58.0 \%$ & 50  & 13.06 & 22.52\\
10 & NA & NA & NA  & $58.8 \%$ &  60  & 17.42 & 29.62\\

\bottomrule

\end{tabular}
}
\vspace{0.2cm}
\captionsetup{font=small,labelfont=bf}
\caption{ Details of the results obtained on CIFAR10 with multiple FL setups. “IID” and “non-IID” terms are employed to distinguish between clients that have an evenly distributed set of samples containing all the classes (IID) and clients that have more samples of certain classes (non-IID). The ``Max Acc.'' column reports the maximum testing accuracy obtained. The ``Rounds'' column gives the number of rounds needed to reach the indicated level of testing accuracy. Finally, ``Carbon Cost'' numbers are obtained by applying Eq. \ref{eq:carboncost} (lower is better).}
\centering
\label{tab:allopti}
\end{table}

\end{document}